\documentclass[10pt,twocolumn,letterpaper]{article}

\usepackage{cvpr}
\usepackage{times}
\usepackage{epsfig}
\usepackage{graphicx}
\usepackage{amsmath}
\usepackage{amssymb}

\usepackage{booktabs}
\usepackage{multirow}
\usepackage[dvipsnames,table]{xcolor}
\usepackage{enumitem}
\usepackage{dsfont}
\usepackage{balance}
\usepackage{placeins}

\usepackage[pagebackref=true,breaklinks=true,colorlinks,citecolor=Green,bookmarks=false]{hyperref}

\cvprfinalcopy 

\def\cvprPaperID{14}

\ifcvprfinal\fi
\begin{document}

\title{Vehicle Re-ID for Surround-view Camera System}

\author{
Zizhang Wu$^1$ \quad Man Wang$^1$ \quad Lingxiao Yin$^1$ \quad Weiwei Sun$^2$ \quad Jason Wang$^1$\quad Huangbin Wu$^3$
\\
$^1$Zongmu Technology \quad $^2$University of Victoria \quad $^3$ Xiamen University
\\
\tt \small \{zizhang.wu, man.wang, linxiao.yin, jason.wang\}@zongmutech.com\\ 
\tt \small \{weiweisun\}@uvic.ca, \{wuhuangbin95\}@stu.xmu.edu.cn
}
\maketitle
\begin{abstract}
     The vehicle re-identification (Re-ID) plays a critical role in the perception system of autonomous driving, which attracts more and more attention in recent years. However, to our best knowledge, there is no existing complete solution for the surround-view system mounted on the vehicle. In this paper, we argue two main challenges in above scenario: i) In single-camera view, it is difficult to recognize the same vehicle from the past image frames due to the fish-eye distortion, occlusion, truncation, etc. ii) In multi-camera view, the appearance of the same vehicle varies greatly from different camera’s viewpoints. Thus, we present an integral vehicle Re-ID solution to address these problems. Specifically, we propose a novel quality evaluation mechanism to balance the effect of tracking box’s drift and target’s consistence. Besides, we take advantage of the Re-ID network based on attention mechanism, then combined with a spatial constraint strategy to further boost the performance between different cameras. The experiments demonstrate that our solution achieves state-of-the-art accuracy while being real-time in practice. Besides, we will release the code and annotated fisheye dataset for the benefit of community.
\end{abstract}

\section{Introduction}
With the advent of autonomous driving, significant efforts have been devoted in the computer vision community to the vehicle-related research. Specially, vehicle Re-ID is one of the most active research fields aiming at identifying the same vehicle across the image archive captured from different cameras. Despite its long success, vehicle Re-ID in real-world scenarios is still a challenging task in computer vision field. As a consequence, it is desirable to seek an effective and robust vehicle Re-ID method, which is the prerequisite for achieving the trajectory prediction, state estimation and speed estimation of the target vehicle.

\begin{figure}[!ht]
    \centering
    \includegraphics[width=1\linewidth]{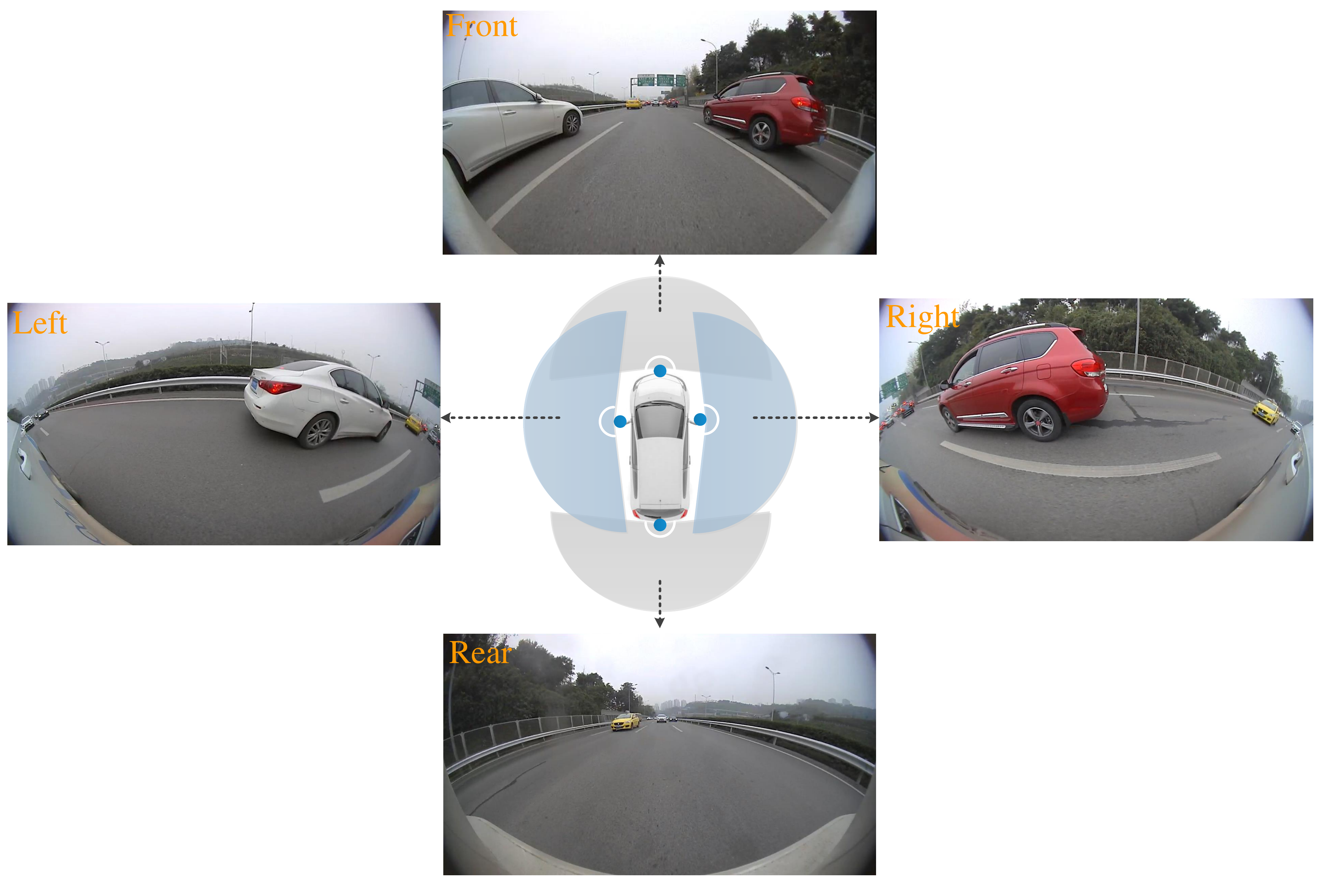}
    \caption{360$^{\circ}$ surround-view camera system. Each arrow points to an image captured by the corresponding camera.}
    \label{fig:Figure01}
\end{figure}

Existing Re-ID studies mainly focus on the pedestrian Re-ID and the vehicle Re-ID. Unlike pedestrian Re-ID \cite{article1}, \cite{article2}, \cite{article3} which extract rich features from images with different poses and colors, vehicle Re-ID faces more severe challenges. Vehicles captured by cameras suffer from distortion, occlusion,  individual similarity, etc. Although the pioneering deep learning-based methods \cite{article14}, \cite{article28}, \cite{article29}, \cite{article30} can learn global semantic features from an entire vehicle image, they still fail to distinguish some small but discriminative regions.
Therefore, \cite{article4}, \cite{article5}, \cite{article6} utilized spatio-temporal relations to learn the similarity between vehicle images, which boost the performance of Re-ID. However, the spatio-temporal information are not annotated in all existing datasets, so there are some restrictions for exploring these methods. To improve the adaptability of vehicle Re-ID methods to different scenarios, \cite{article7}, \cite{article8}, \cite{article31} employ vehicle attributes to learn the correlations among vehicles. Although the vehicle attributes are more general and discriminating than other features, the relationship between vehicle attributes and categories is ignored. Subsequently, \cite{article9} introduced the attention mechanism in the attribute branch. This operation is conducive to select attributes for corresponding input vehicle images, which helps the category branch select better discriminative features for category recognition.

Despite tremendous progresses made in recent vehicle Re-ID methods, most of them are designed for surveillance scenario. While in the field of autonomous driving, surround-view camera systems become more and more popular. To achieve a 360$^{\circ}$ blind-spot free environment perception, we mount multiple fisheye cameras around the vehicle, as shown in Fig.\ref{fig:Figure01}. Therefore, it is essential to learn the relationship between the same vehicle among different vehicle-mounted cameras and among different frames in single camera. We state two main challenges for achieving a vehicle Re-ID solution in surround-view multi-camera scenario: 

1) In single-camera view, vehicle features in consecutive frames vary dramatically due to fish-eye distortion, occlusion, truncation, etc. It is difficult to recognize the same vehicle from the past image archive under such interference.

2) In multi-camera view, the appearance of the same vehicle varies significantly from different viewpoints. Individual similarity of vehicles also led to great confusion for matching. 

In this paper, we propose our methods respectively to these challenges. For the balance of accuracy and efficiency,  SiamRPN++ network \cite{article10} and BDBnet \cite{article11} are employed for vehicle  re-identification in single camera and multi-camera, respectively. However, these two models are mainly designed for surveillance scenes, and are unable to deal with the significant variations of vehicle's appearance in fisheye system. Therefore, a post-process module named quality evaluation mechanism for the output bounding box is proposed to alleviate the target drift caused by fish-eye distortion, occlusion, etc. Besides, an attention module and a spatial constraint strategy are introduced to respond the intra-class difference and inter-class similarity of vehicles \cite{article7}. To drive the study of vehicle Re-ID in surround-view scenarios and fill the gap in relative dataset, we will release the large-scale annotated fisheye dataset.

Our contributions can be summarized as follows:

\begin{itemize}

\item We design an integral vehicle Re-ID solution for the surround-view multi-camera scenario.
\item We propose a quality evaluation mechanism for the output bounding box, which can alleviate the problem of target drift. 
\item We utilize novel spatial constraint strategy for regularizing the Re-ID results in the surround-view camera system.
\item We will release a large-scale fisheye dataset with annotations to promote the relevant research.
\end{itemize}

\section{Related Work}

In this section, we briefly review the literatures of Vehicle Re-Id methods and the tracking algorithm, which are both closely related to our works.

\paragraph{Vehicle Re-ID}:  Vehicle Re-ID has been widely studied in recent years. As well-known, the common challenge is how to deal with the inter-class similarity and the intra-class difference \cite{article7} -- Different vehicles have similar appearance, while the same vehicle looks different due to the diverse perspectives and distortion. Until now, lots of works have been proposed to tackle this challenge. Liu et al. \cite{article7} designed a pipeline, which adopts deep relative distance learning (DRDL) to project vehicle images into Euclidean space, then calculate the distance in Euclidean space to measure the similarity of two vehicle images. Liu et al. \cite{article4} created a dataset named VeRi-776, which employs visual features, license plate and spatial-temporal information to explore the Re-ID task of vehicles. Meanwhile, further works \cite{article5}, \cite{article6}, \cite{article12}, \cite{article33}, \cite{article32} introduced spatial-temporal information to boost the performance of Re-ID . Shen et al. \cite{article5} proposed a two-stage framework, which utilizes complex spatial-temporal information of vehicles to effectively regularize Re-ID results. Subsequently, \cite{article6} used spatial prior knowledge to generate the tracklet and selected the vehicle in the middle frame as a feature of tracklet. Furthermore, \cite{article12} utilized location information between cameras to improve the accuracy of Re-ID. In the light of success of using location information, we exploit a novel spatial constraint strategy to enhance the Re-ID results in surround-view camera system. 

Since the spatial-temporal information in datasets is not always available, approaches for ReId have been proposed by combining local and global features of targets \cite{article13}, \cite{article11}, \cite{article14}, \cite{article15}, \cite{article34}, \cite{article35}. For instances, \cite{article13} designed a network with the combination of local and global branch, which employs joint feature vectors to enhance robustness on occlusion problems. Different from the DropBlock, \cite{article11} proposed Batch DropBlock that drops the same region in a batch of image to better accomplish the metric learning task. \cite{article14} enabled the model to perform intensive learning of local features from the loss function aspect. Inspired by the attention mechanism, methods of processing the local and global information are more flexible \cite{article36}, \cite{article37}, \cite{article38}, \cite{article39}, \cite{article40}. In this paper, we also utilize the attention mechanism to make the model focus on target regions, and use triplet loss \cite{article15} and softmax loss to enhance the performance of vehicle Re-ID. 

\paragraph{Tracking algorithms related with Vehicle Re-ID:}  Object tracking algorithm plays an important role in the implementation of vehicle Re-ID scheme \cite{article16}, \cite{article17}, \cite{article18}, \cite{article19}. For object tracking tasks, the trackers based on Siamese network \cite{article20},\cite{article21},\cite{article22},\cite{article23},\cite{article24} have received signiﬁcant attentions for their well-balanced tracking accuracy and efﬁciency, which is of great value for boosting vehicle Re-ID task in the surround-view multi-camera scenario. The seminal work \cite{article20}, \cite{article21}, \cite{article22} formulated visual tracking as a cross-correlation problem and produced a tracking similarity map from the depth model with Siamese network, so as to find the location of the target by comparing the similarity between the target and the search region. However, these works have an inherent drawback, as their tracking accuracies on the OTB benchmark \cite{article25} still leave a relatively large gap with state-of-the-art deep trackers like ECO \cite{article26} and MDNet \cite{article27}. To overcome this drawback, Li et al. \cite{article10} transferred the object position in the bounding box during the training phase to avoid the location bias of the network. Thus, the network is able to focus on the object marginal area of the search region and the accuracy of Siamese trackers is boosted signiﬁcantly. In addition, this approach also proposed a lightweight depth-wise cross-correlation layer to improve the running speed. Considering its great performance, we adopt the SiamRPN++ model to realize the vehicle Re-ID task in single camera.

\section{Vehicle Re-ID in surround-view camera system}
In the structure of surround-view multi-camera system, we divide vehicle Re-ID task into two subtasks: Re-ID process in single camera and in multi-camera. In this work, we explain our methods within each subtask.

\subsection{Vehicle Re-ID in single camera}

\begin{figure}
		\begin{minipage}{1\linewidth}
			\centering
			\centerline{\includegraphics[width=8.5cm]{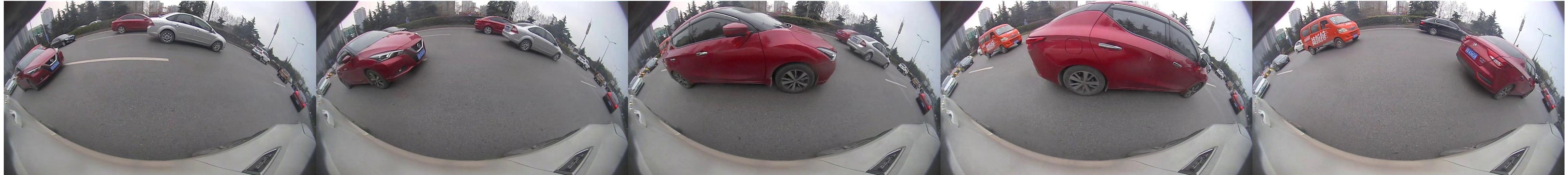}}
			\centerline{(a)}
		\end{minipage}
		
		\begin{minipage}{1\linewidth}
			\centering
			\centerline{\includegraphics[width=8.5cm]{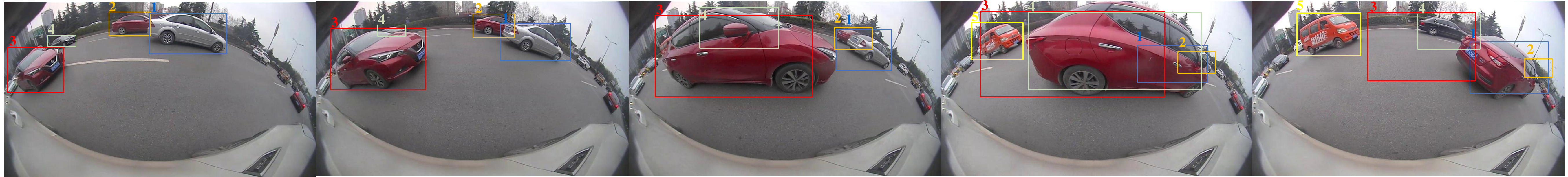}}
			\centerline{(b)}
		\end{minipage}
		
		\begin{minipage}{1\linewidth}
			\centering
			\centerline{\includegraphics[width=8.5cm]{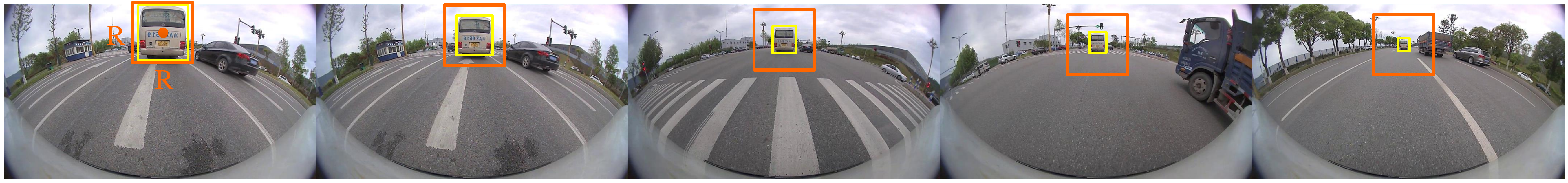}}
			\centerline{(c)}
		\end{minipage}
	
        \caption{Vehicles in single view of fisheye camera. (a) The same vehicle features change dramatically in consecutive frames and vehicles tend to obscure each other. (b) Matching errors are caused by tracking results. (c) The vehicle center indicated by the orange box is stable  while the IoU in consecutive frames indicated by the yellow box decreases with movement.}
        \label{fig:Figure02}
	\end{figure}

Single camera vehicle Re-ID task aims at matching vehicles from the same view in consecutive frames. We utilize SiamRPN++ \cite{article10} as the single tracker model and place such tracker for each target to realize Re-ID in a single camera. Despite the great success of SiamRPN++, we observe that it still fails when the target suffers from large distortion rate in different positions of the camera, which enlarges differences between target features in different frames. Besides, occlusion between targets leads to more complexity of target location information. Failure cases are shown in Fig. \ref{fig:Figure02}(b). To circumvent challenges above, we propose the novel post-process method for data association as follows.



\begin{figure*}[!ht]
    \centering
    \includegraphics[width=16.5cm]{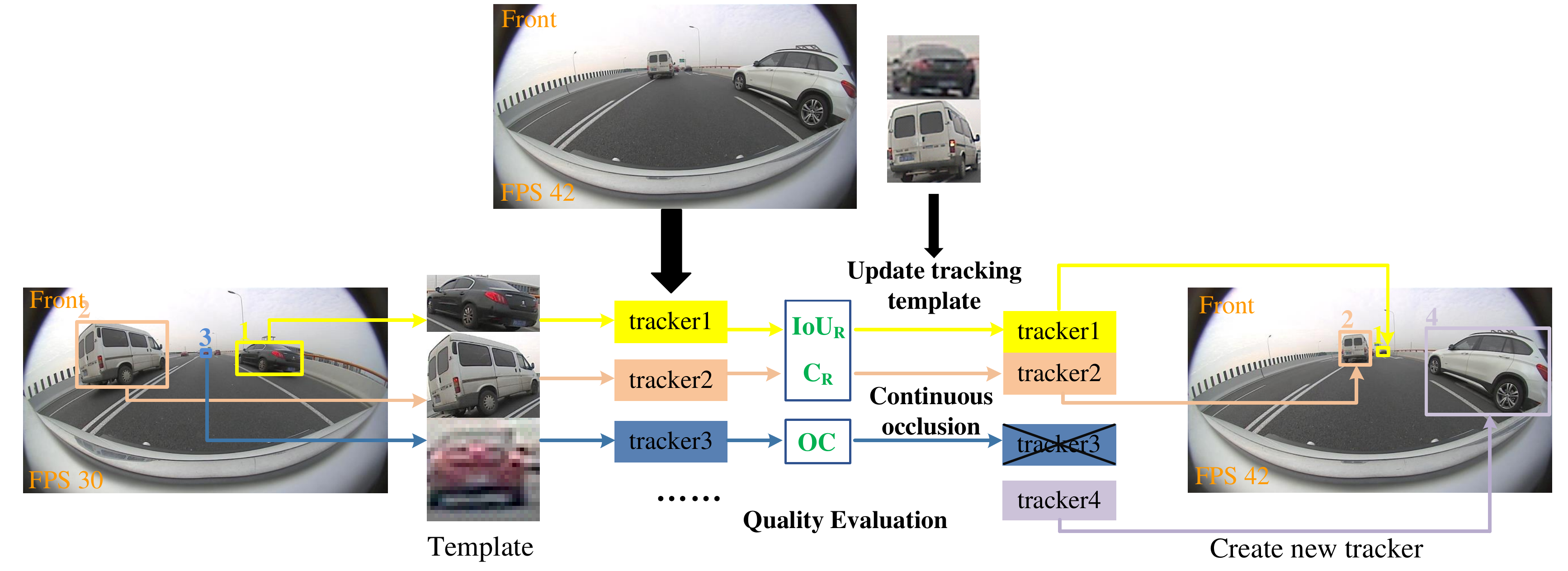}
    \caption{The overall framework of vehicle Re-ID in single camera. Each object is assigned a single tracker to realize Re-ID in single channel. Tracking templates are initialized with object detection results. All tracking outputs are post-processed by the quality evaluation module to deal with the distorted or occluded objects.}
    \label{fig:Figure04}
\end{figure*}

\subsubsection{Quality Evaluation Mechanism.} 
Unlike tracking task, vehicle Re-ID task has fewer restrictions on the size of the bounding box, but instead it is sensitive to the drift level of the bounding box center. Therefore, the post-processing methods of tracking results must focus on the requirements of Re-ID task. Inspired by the attention mechanism, we propose a novel quality evaluation mechanism to optimize the tracking results.

\textbf{Center Drift.} It is essential to update the tracking template to adapt target movement. $IoU$ and confidence of output bounding box in consecutive frames are usually taken as indicators to dynamically update the template. Comparatively, Re-ID process pays more attention to center drift of the target. In Fig. \ref{fig:Figure02}(b), severe center drift results in much matching errors. However, IoU is not an appropriate metric for the task here. In Fig. \ref{fig:Figure02}(c), the center of target vehicle is stable while IoU decreases continuously because of the changing size of the bounding box. Updating templates in such circumstance may consume more resources and have higher risk of wrong predictions. To alleviate this problem, we define a center drift metric, $IoU_R$, to measure the drift level of the target center.

\begin{equation}
IoU_R=\frac{S}{2R^2-S},
\label{Eq(1)}
\end{equation}
where $R$ is side length of the orange square in the output bounding box center as shown in Fig.  \ref{fig:Figure02}(d), we set it as a constant in experiments. $S$ is the intersection area between tracking results of the same target in consecutive frames.

\textbf{Re-ID Confidence.} The confidence score output from tracking process are used for ranking the bounding boxes. It is closely related with the center position and size of bounding box, which is improper for Re-ID processes. Therefore, we suggest Re-ID confidence ($C_R$) to verify accuracy of Re-ID result as Eq. (\ref{Eq(2)}).
\begin{equation}
C_R=C_T \times IoU_R,
\label{Eq(2)}
\end{equation}
where $C_T$ is the tracking confidence. The drift level can downweight the scores of bounding boxes far from the previous center of an object. 

We define the conditions for updating the tracking template based on $IoU_R$ and $C_R$ as follows:

\begin{equation}
IoU_{RM}<T_1, C_{RM}<T_2,
\label{Eq(3)}
\end{equation}
where $M$ represents the average result of consecutive $M$ frames, $T_1$,$T_2$ are corresponding thresholds for $IoU_R$ and $C_R$. The template adaptive updating process adapted to the process of Re-ID task is realized. 

\textbf{Occlusion Coefficient}
To alleviate the box drift caused by occlusion, we introduce the occlusion coefficient to quantify the degree of occlusion as Eq. (\ref{Eq(4)}):
\begin{equation}
OC=\frac{I_N}{A},
\label{Eq(4)}
\end{equation}
where $I_N$ stands for the intersection area between two tracking results of objects in the same frame, $A$ is the area of the object. An object is defined as severely occluded when $OC$ is greater than the threshold $T_O$. When both objects have high overlapping rates, the object with lower $C_R$ is counted as occluded. We take consecutive frames results as the criterion to deal with the occluded target as the position relation changes with time. The tracker and ID of the obscured object will be maintained until $N$ consecutive frames of occlusion, otherwise they would be removed forever.

The $IoU_R$, $C_R$ and $OC$ constitute the quality evaluation mechanism, which processes the tracking results and optimizes the Re-ID performance in single camera.

\subsubsection{Framework of vehicle Re-ID in single camera}
The overall framework of vehicle Re-ID in single camera is depicted as Fig. \ref{fig:Figure04}. Each object is assigned a single tracker whose template is initialized with object detection result. These trackers deal with the next frame and output Re-ID results to the quality evaluation module. An unqualified result leads to template update and another tracking process. The tracker and ID of occluded objects will be deleted or maintained temporarily depending on the number of consecutive frames.

\subsection{Vehicle Re-ID in multi-camera}
\begin{figure}
		\begin{minipage}{1\linewidth}
			\centering
			\centerline{\includegraphics[width=8cm]{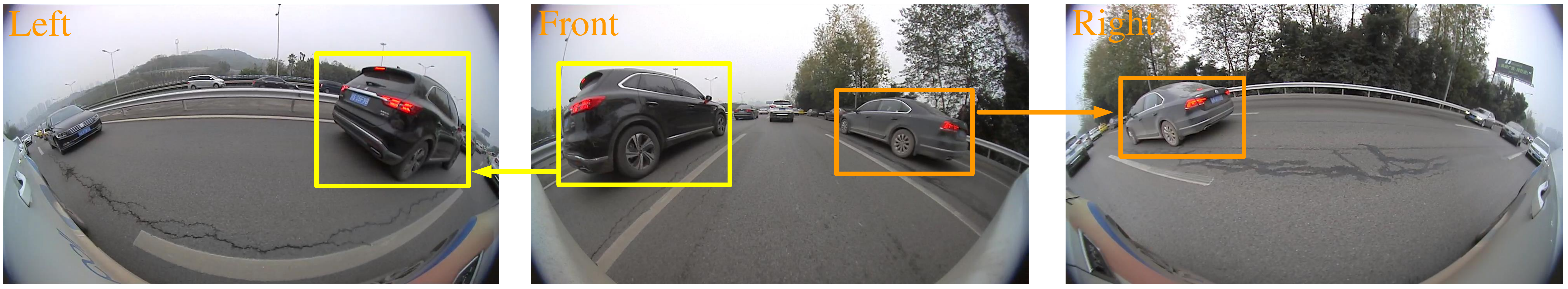}}
			\centerline{(a)}
		\end{minipage}
		
		\begin{minipage}{1\linewidth}
			\centering
			\centerline{\includegraphics[width=8cm]{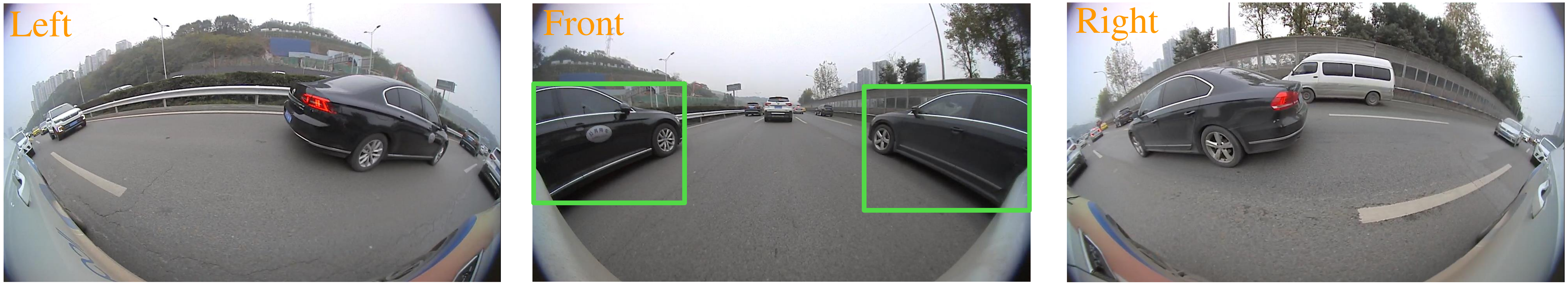}}
			\centerline{(b)}
		\end{minipage}
		
        \caption{Samples captured by different cameras. (a) The appearances of the same vehicle captured by different cameras vary greatly, and the same color represents the same object. (b) Objects have a similar appearance may appear in the same camera view, as shown by these two black vehicles in green boxes.}
        \label{fig:Figure05}
	\end{figure}

Vehicle Re-ID in multi-camera aims at building correlation between the same vehicle in different cameras. Most of the current methods employ deep network to achieve Re-ID. However, in surround view camera system, cameras are mounted at different positions around the vehicle, resulting in the same object appears variously in different cameras, as shown in Fig. \ref{fig:Figure05}(a). Understanding the fact that adopting general deep learning networks is incapable of handing this problem. Therefore, we introduce an attention module in this paper that forces the network to pay more attention to target areas. Furthermore, different targets appear in the same camera may have similar appearances, as shown in Fig. \ref{fig:Figure05}(b). It is difficult to distinguish these two black vehicles in front camera only use the features in image-level. Consequently, we introduce a novel spatial constraint strategy to handle this thorny problem.

\subsubsection{Attention Module}
\begin{figure}[!ht]
    \centering
    \includegraphics[width=7.5cm]{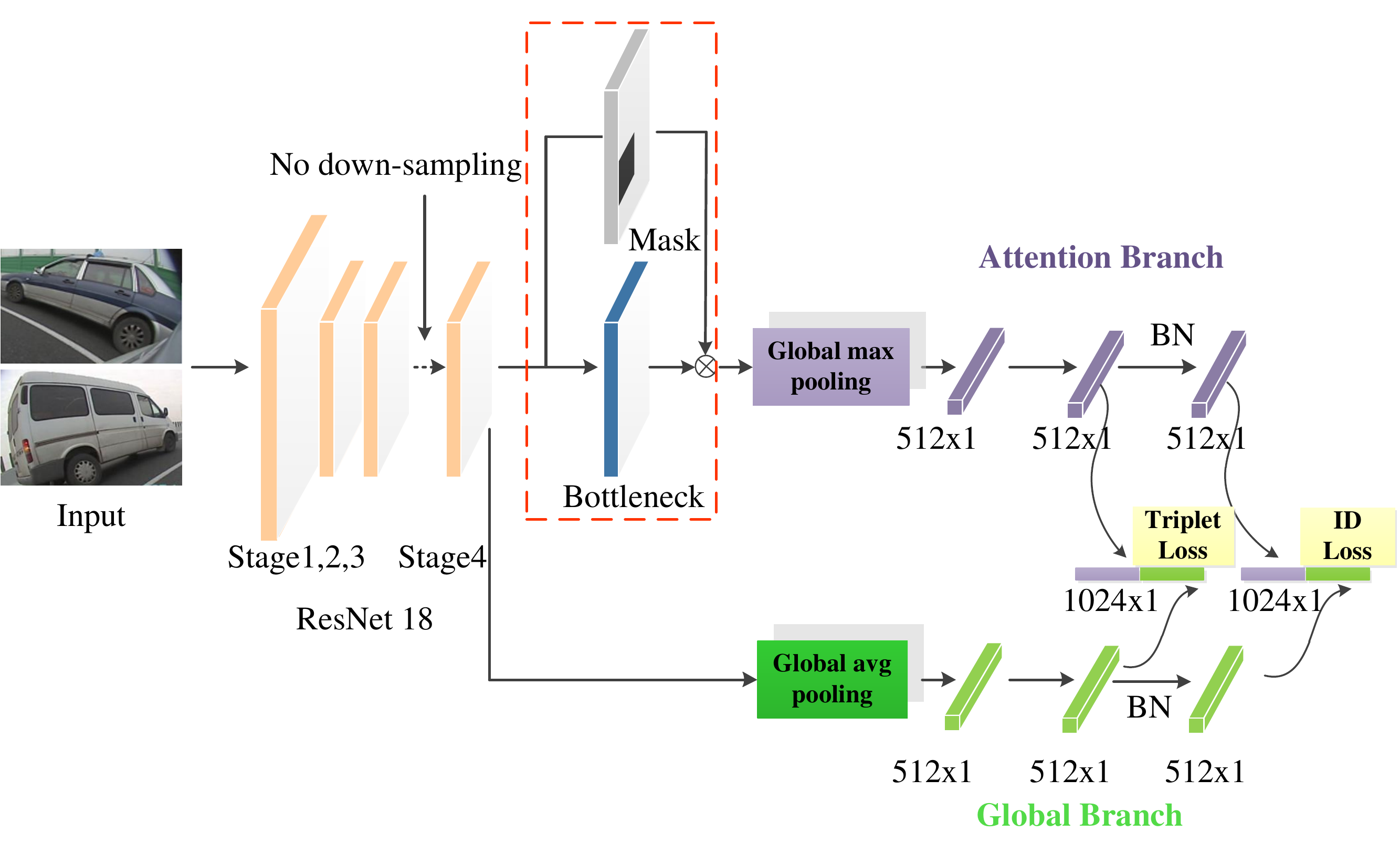}
    \caption{Illustration of the multi-camera Re-ID network. This network is a two branch parallel structure. The top branch is employed to make the network pay more attention on object regions, and anther is for extracting global features.}
    \label{fig:Figure06}
\end{figure}

Existing vehicle Re-ID methods mainly serve for surveillance scenarios. To meet the requirement of vehicle Re-ID for surround-view camera system, we apply a modified BDBnet \cite{article11} as our multi-camera Re-ID model. BDBnet consists of a global branch and a local branch. Specially, a fixed mask is added to local branch to help the network learn semantic features, and it is shown to be effective in pedestrian Re-ID. Different from pedestrian Re-ID, vehicle Re-ID for surround-view camera system suffers from deformation in multi-camera system. Fixed templates are difficult to improve learning outcome, so we introduce an attention module that leads the network to learn self-adaptive template to focus on target regions. As shown in Fig. \ref{fig:Figure06}, the structure in red box is the attention module. For each new target, the network is applied to extract features and measure Euclidean distance between this feature and features stored in feature gallery. Then the distance is converted to confidence score $s_1$ through Eq. (\ref{Eq(5)}). 

\begin{equation}
s_1=ln(\frac{1}{D_F}+1),
\label{Eq(5)}
\end{equation}
\noindent
where $D_F$ is the Euclidean distance between the feature of the new target and features stored in gallery.

\subsubsection{Spatial Constraint Strategy}
\begin{figure}[!ht]
    \centering
    \includegraphics[width=8.0cm]{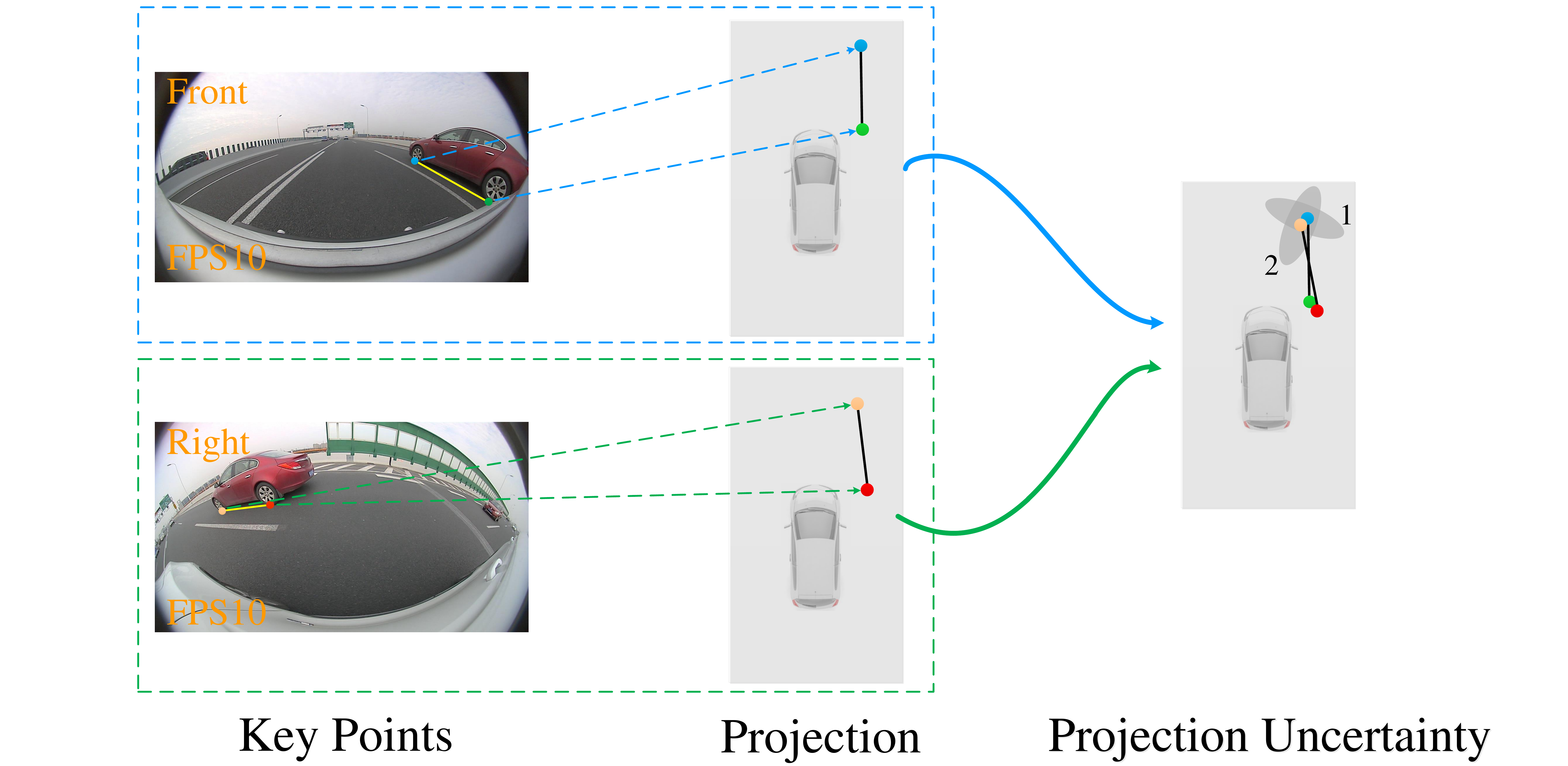}
    \caption{Projection uncertainty of key points.
    Ellipse 1 and ellipse 2 are uncertainty ranges of front and left (right) cameras, respectively. }
    \label{fig:Figure07}
\end{figure}

To deal with scenarios containing multiple similar targets in single camera, we introduce a novel spatial constraint strategy. As we know, the wheel grounding key points of the same target in different cameras correspond to the same position in real world coordinates. However, the projected position varies due to external factors, such as camera mounted angle and calibration. We define the offset caused by these factors as projection uncertainty as shown in Fig. \ref{fig:Figure07}. Two key points of the same category which are projected into the overlapping area are determined to belong to the same vehicle. Furthermore, error is decreased if target gets closer to the camera, so we suggest different standard for wheels in different position. As presented in Eq. (\ref{Eq(6)}), we first calculate coordinate distances between key points and then convert them to score $s_2$:
\begin{equation}
s_2=ln(\frac{1}{D_K}+1),
\label{Eq(6)}
\end{equation}
where $D_K$ is the distance between the projection coordinate of the key points, and $D_K=D_f+D_r$, $D_f$ and $D_r$ are the projection coordinate distance of these two front key points and two rear key points, respectively.

\begin{figure*}[!ht]
    \centering
    \includegraphics[width=15cm]{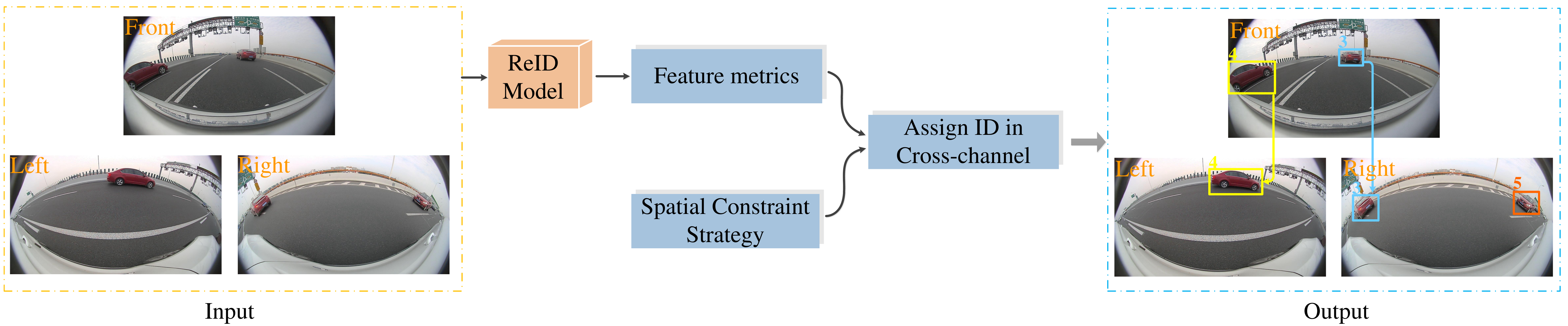}
    \caption{The overall framework of the vehicle Re-ID in multi-camera.For the new target, Re-ID model is used first to extract the features,followed by the distance metrics is carried out for this feature and features in gallery. Besides, the spatial constraint strategy is adopted to improve the correlation effect.}
    \label{fig:Figure08}
\end{figure*}
\subsubsection{Framework of vehicle Re-ID in multi-camera}
The integral process of multi-camera vehicle Re-ID is shown in Fig.\ref{fig:Figure08}. The first branch is used to obtain the confidence score $s_1$ of feature similarity metrics, and the second branch is used to obtain the confidence score $s_2$ of physical coordinate distances of key points. Finally, ID of the target with the highest score $s$ is assigned to the new target.
\begin{equation}
s=\frac{\alpha s_1+\beta s_2}{\alpha+\beta},
\label{Eq(7)}
\end{equation}
where $\alpha$ and $\beta$ are set to 1 in following experiments.

The multi-camera ID association in this paper is implemented in a certain order. The association between left camera and front camera is carried out first, followed by the association between front camera and right camera. In detail, the process is as follows:

1)	For the case that a new target appears on the left side of the left camera or the right side of the right camera, the association only needs to be carried out in their own camera. If successful, we assigned the original ID for the new target. If not, we assign a new ID for this target immediately.

2)	When a new target appears on other areas of the left or right camera, it just needs to be associated with the front camera. 

3)	When a new target appears on right side of the front camera, only the right camera needs to be associated. Similarly, it just needs to be associated with the left camera for the new target appears on the left side of the front camera.

\subsection{Overall framework}
The surround-view multi-camera Re-ID system processes data of each camera serially. Each new object in single channel will be assigned a single target tracker and matched with data of other channels according to the Re-ID strategy in multi-camera. If not matched successfully, a new ID will be created for it. Else, the ID of the matching object will be inherited by the newcomer. All targets in single camera will be matched in time sequence according to the Re-ID strategy in single camera.

\section{Experiments}
\noindent
\subsection{Dataset and evaluation metric}

We generate the fisheye dataset from structured roads. It consists of a total of 22190 annotated images sampled from 36 sequences. The resolution of image is 1280$\times$720. We use the 80$\%$ and 20$\%$ images as training and testing set. 
We utilize three cameras (i.e., front, left and right) to capture these images at 30 frames per second (fps). 

We evaluate the results with the identity consistency ($IC$). It is formulated as 
\begin{equation}
IC=1-\frac{\sum_tIDSW_t}{\sum_tID_t},
\label{Eq(8)}
\end{equation}
\noindent
where $t$ is the frame index, identity switch (IDSW) is counted if a ground truth target $i$ is matched to tracking output $j$ and the last known assignment is $k$, $k\neq j$. $ID_t$ is the sum of ground truth targets with ID in frame $t$.

\subsection{Implementation Details}
We train all networks with stochastic gradient descent (SGD) on GTX 1080Ti.
For the single camera Re-ID model SiamRPN++, the number of training epochs is 50. Batch size is set as 24 on training stage due to the limit of the GPU memory. Learning rate is 0.0005. Weight decay is 0.0001. Momentum is 0.9. In the case of multi-camera Re-ID model, the weight decay and momentum are set as 0.9 and 0.0005 respectively. We use an initial learning rate of 0.00035 for the first 10 epochs. We train the network with 150 epochs. Batch size is 256.

\subsection{Evaluation of the proposed method}
\textbf{Quality Evaluation Mechanism.} The proposed quality evaluation mechanism is a key component to optimize Re-ID performance in single camera. We conduct some ablation studies on our dataset to find out the contribution of our method to performance. 

We first compare the template updating metrics. As shown in Table \ref{table01}, Default is our baseline model without updating templates. It is obvious that updating the tracking template with $IoU_T$ and $C_T$ as metrics brings significant improvement on identity consistency. Furthermore, when utilizing the $IoU_R$ and $C_R$, the identity consistency is greatly improved once again. That means the revised metrics helps the model pay more attention to demands of the Re-ID task.

\begin{table}[h]
\centering
\setlength{\abovecaptionskip}{2pt}%
\setlength{\belowcaptionskip}{2pt}%

\begin{tabular}{@{}lccc@{}}
\toprule
               & $IC_{front}$ & $IC_{left}$ & $IC_{right}$     \\ \midrule
Default         & 0.82      & 0.81     & 0.87      \\
+$IoU_T+C_T$    & 0.88      & 0.91     & 0.90      \\
+$IoU_R+C_R$    & 0.94      & 0.96     & 0.97      \\

\bottomrule
\end{tabular}
\caption{The impact of the different methods updating tracking template. $IC_{front}$, $IC_{left}$ and $IC_{right}$ correspond to the $IC$ of front, left and right cameras }\label{table01}
\end{table}

The number of frames ($N$) to delete the occluded object is summarized as Table \ref{table02}. Zero frame means that we do not handle the occluded cars. When $N=2$, the identity switch is decrease slightly. It is expected because deleting the temporary occluded cars frequently contributing the $ID$ switch as its rapid reappearance. With $N$ grows, the $ID$ is gradually steady. However, maintaining too much frames means don’t handle occlusion and the consistency descends once again. Experimental results show that $N = 4$ is optimal in this paper.

\begin{table}[h]
\centering
\setlength{\abovecaptionskip}{2pt}%
\setlength{\belowcaptionskip}{2pt}%

\begin{tabular}{@{}lccc@{}}
\toprule
N   & $IC_{front}$ & $IC_{left}$    & $IC_{right}$          \\ \midrule
0   & 0.83      & 0.81     & 0.80      \\
2   & 0.85      & 0.83     & 0.84      \\
3   & 0.92      & 0.90     & 0.93      \\
4   & 0.94      & 0.96     & 0.97      \\
5   & 0.93      & 0.91     & 0.92      \\
\bottomrule
\end{tabular}
\caption{The impact of the different methods updating tracking template. }\label{table02}
\end{table}

\textbf{Vehicle Re-ID strategy in multi-camera.} We examine the influence of various matching strategies in multi-camera in Table \ref{table03}. All the experiments are based on the same implementations in single view. The first row shows the method of matching with feature metrics as \cite{article11}. After introducing the attention module, the Re-ID accuracy is improved a lot. Base on that, adding the spatial constraints strategy provide a nice improvement as show in the last row. 


\begin{table}[h]
\centering
\setlength{\abovecaptionskip}{2pt}%
\setlength{\belowcaptionskip}{2pt}%

\resizebox{\linewidth}{!}{
\begin{tabular}{@{}lccc@{}}
\toprule
ID Matching strategy in multi-camera  & $IC_{front}$  & $IC_{left}$    & $IC_{right}$      \\ \midrule
Baseline                      & 0.84      & 0.85     & 0.86      \\
+Attention module             & 0.87      & 0.92     & 0.89      \\
\begin{tabular}[l]{@{}l@{}}+Attention module +Spatial \\  constraint strategy\end{tabular} & 0.94      & 0.96     & 0.97 \\

\bottomrule
\end{tabular}}
\caption{The impact of the different methods updating tracking template. }\label{table03}
\end{table}

\section{Conclusion}
In this paper, we propose a complete vehicle Re-ID solution for the surround-view multi-camera scenario. The introduced quality evaluation mechanism for output bounding box can alleviate the problems caused by distortion and occlusion. Besides, we deploy attention module to make the network focus on target regions. A novel spatial constraint strategy is also introduced to significantly regularize the Re-ID results in this scenario. Extensive component analysis and comparison experiments on the fisheye dataset show that our vehicle Re-ID solution achieves promising results. Furthermore, the annotated fisheye dataset will be public to advance the development of the research on this field. In future studies, we will further optimize the performance of vehicle Re-ID in the surround-view multi-camera scenario.

{\small
\bibliographystyle{ieee_fullname}
\bibliography{egbib}
}

\end{document}